\documentclass[11pt]{article}
\pdfoutput=1
\usepackage{fullpage}
\usepackage{graphicx}
\usepackage{tabularx}
\usepackage{amssymb}
\usepackage{xcolor}
\usepackage[normalem]{ulem}
\usepackage{amsmath}
\usepackage{authblk}
\usepackage{url}
\usepackage{multirow}
\usepackage{booktabs}
\usepackage{adjustbox}

\usepackage{amsmath}
\usepackage{textcomp}
\usepackage{amssymb}
\usepackage[utf8]{inputenc}
\usepackage[textwidth=1.cm, textsize=tiny]{todonotes}
\usepackage{multirow}
\usepackage{url}

\usepackage{cite}
\usepackage{soul,xcolor}
\usepackage{amsmath}
\usepackage{array}

\begin{document}
\setstcolor{red}
\title{\textbf{EEGNet: A Compact Convolutional Neural Network for EEG-based Brain-Computer Interfaces}}

\author[1,*]{Vernon J. Lawhern}
\author[1,2]{Amelia J. Solon}
\author[1,3]{Nicholas R. Waytowich}
\author[1,2]{Stephen M. Gordon}
\author[1,4]{Chou P. Hung}
\author[1]{Brent J. Lance}

\affil[1]{Human Research and Engineering Directorate, U.S. Army Research Laboratory, Aberdeen Proving Ground, MD}
\affil[2]{DCS Corporation, Alexandria, VA}
\affil[3]{Department of Biomedical Engineering, Columbia University, New York, NY}
\affil[4]{Department of Neuroscience, Georgetown University, Washington, DC}
\affil[*]{Corresponding Author}

\maketitle

\begin{abstract}
\textit{Objective}: Brain computer interfaces (BCI) enable direct communication with a computer, using neural activity as the control signal. This neural signal is generally chosen from a variety of well-studied electroencephalogram (EEG) signals. For a given BCI paradigm, feature extractors and classifiers are tailored to the distinct characteristics of its expected EEG control signal, limiting its application to that specific signal. Convolutional Neural Networks (CNNs), which have been used in computer vision and speech recognition to perform automatic feature extraction and classification, have successfully been applied to EEG-based BCIs; however, they have mainly been applied to single BCI paradigms and thus it remains unclear how these architectures generalize to other paradigms. Here, we ask if we can design a single CNN architecture to accurately classify EEG signals from different BCI paradigms, while simultaneously being as compact as possible (defined as the number of parameters in the model). \textit{Approach}: In this work we introduce EEGNet, a compact convolutional neural network for EEG-based BCIs. We introduce the use of depthwise and separable convolutions to construct an EEG-specific model which encapsulates well-known EEG feature extraction concepts for BCI. We compare EEGNet, both for within-subject and cross-subject classification, to current state-of-the-art approaches across four BCI paradigms: P300 visual-evoked potentials, error-related negativity responses (ERN), movement-related cortical potentials (MRCP), and sensory motor rhythms (SMR). \textit{Results}: We show that EEGNet generalizes across paradigms better than, and achieves comparably high performance to, the reference algorithms when only limited training data is available. We also show that EEGNet effectively generalizes to both ERP and oscillatory-based BCIs. In addition, we demonstrate three different approaches to visualize the contents of a trained EEGNet model to enable interpretation of the learned features. \textit{Significance}: Our results suggest that EEGNet is robust enough to learn a wide variety of interpretable features over a range of BCI tasks, suggesting that the observed performances were not due to artifact or noise sources in the data. Our models can be found at: https://github.com/vlawhern/arl-eegmodels.

\end{abstract}


\section{Introduction}\label{sec:introduction}

A Brain-Computer Interface (BCI) enables direct communication with a machine via brain signals \cite{Wolpaw2002}. Traditionally, BCIs have been used for medical applications such as neural control of prosthetic artificial limbs \cite{Schwartz2006}.  However, recent research has opened up the possibility for novel BCIs focused on enhancing performance of healthy users, often with noninvasive approaches based on electroencephalography (EEG) \cite{vanerp_braincomputer_2012, Saproo2016, Lance2012}. Generally speaking, a BCI consists of five main processing stages \cite{NicholasAlonso2012}: a data collection stage, where neural data is recorded; a signal processing stage, where the recorded data is preprocessed and cleaned; a feature extraction stage, where meaningful information is extracted from the neural data; a classification stage, where a decision is interpreted from the data; and a feedback stage, where the result of that decision is provided to the user. While these stages are largely the same across BCI paradigms, each paradigm relies on manual specification of signal processing \cite{Bashashati2007}, feature extraction \cite{McFarland2006} and classification methods \cite{Lotte2007}, a process which often requires significant subject-matter expertise and/or \textit{a priori} knowledge about the expected EEG signal. It is also possible that, because the EEG signal preprocessing steps are often very specific to the EEG feature of interest (for example, band-pass filtering to a specific frequency range), that other potentially relevant EEG features could be excluded from analysis (for example, features outside of the band-pass frequency range). The need for robust feature extraction techniques will only continue to increase as BCI technologies evolve into new application domains \cite{Zander2011, Lance2012, Saproo2016, vanerp_braincomputer_2012, Blankertz2010, Gordon2017}. 

\textit{Deep Learning} has largely alleviated the need for manual feature extraction, achieving state-of-the-art performance in fields such as computer vision and speech recognition \cite{Hinton2012, LeCun2015}. Specifically, the use of deep convolutional neural networks (CNNs) has grown due in part to their success in many challenging image classification problems \cite{Krizhevsky2012, SimonyanZ14a, Szegedy2014_Inception, He2015, HuangLW16a}, surpassing methods relying on hand-crafted features (see \cite{LeCun2015} and \cite{Schmidhuber2014} for recent reviews). Although the majority of BCI systems still rely on the use of handcrafted features, many recent works have explored the application of Deep Learning to EEG signals. For example, CNNs have been used for epilepsy prediction and monitoring \cite{Antoniades2016, Liang2016, Page2016, Mirowski20091927, ThodoroffPL16}, for auditory music retrieval \cite{Stober2014, StoberSOG15}, for detection of visual-evoked responses \cite{Cecotti2011, Manor2015, Shamwell2016, Cecotti2014} and for motor imagery classification \cite{Schirrmeister2017b}, while Deep Belief Networks (DBNs) have been used for sleep stage detection \cite{Langkvist2012}, anomaly detection \cite{Wulsin2011} and in motion-onset visual-evoked potential classification \cite{Ma2017}. CNNs using time-frequency transforms of EEG data were used for mental workload classification \cite{Bashivan2015} and for motor imagery classification \cite{Tabar2017, An2014, Sakhavi2015}). Restricted Boltzman Machines (RBMs) have been used for motor imagery \cite{Lu2017}. An adaptive method based on stacked denoising autoencoders has been proposed for mental workload classification \cite{Yin2017b}). These studies focused primarily on classification in a single BCI task, often times using task-specific knowledge in designing the network architecture. In addition, the amount of data used to train these networks varied significantly across studies, in part due to the difficulty in collecting data under different experimental designs. Thus, it remains unclear how these previous deep learning approaches would generalize both to other BCI tasks as well as to variable training data sizes.

In this work we introduce \textit{EEGNet}, a compact CNN for classification and interpretation of EEG-based BCIs. We introduce the use of \textit{Depthwise} and \textit{Separable} convolutions, previously used in computer vision \cite{Chollet16a}, to construct an EEG-specific network that encapsulates several well-known EEG feature extraction concepts, such as optimal spatial filtering and filter-bank construction, while simultaneously reducing the number of trainable parameters to fit when compared to existing approaches. We evaluate the generalizability of EEGNet on EEG datasets collected from four different BCI paradigms: P300 visual-evoked potential (P300), error-related negativity (ERN), movement-related cortical potential (MRCP) and the sensory motor rhythm (SMR), representing a spectrum of paradigms based on classification of Event-Related Potentials (P300, ERN, MRCP) as well as classification of oscillatory components (SMR). In addition, each of these data collections contained varying amounts of data, allowing us to explore the efficacy of EEGNet on various training data sizes. Our results are as follows: We show that EEGNet achieves improved classification performance over an existing paradigm-agnostic EEG CNN model across nearly all tested paradigms when limited training data is available. In addition, we show that EEGNet effectively generalizes across all tested paradigms. We also show that EEGNet performs just as well as a more paradigm-specific EEG CNN model, but with two orders of magnitude fewer parameters to fit, representing a more efficient use of model parameters (an aspect that has been explored in previous deep learning literature, see \cite{Chollet16a,Yang2015}). Finally, through the use of feature visualization and model ablation analysis, we show that neurophysiologically interpretable features can be extracted from the EEGNet model. This is important as CNNs, despite their ability for robust and automatic feature extraction, often produce hard to interpret features. For neuroscience practitioners, the ability to derive insights into CNN-derived neurophysiological phenomena may be just as important as achieving good classification performance, depending on the intended application. We validate our architecture's ability to extract neurophysiologically interpretable signals on several well-studied BCI paradigms to show that the network performance is not being driven by noise or artifact signals in the data.

The remainder of this manuscript is structured as follows. Section \ref{datasets} gives a brief description of the four datasets used to validate our CNN model. Section \ref{classification-methods} describes our EEGNet model as well as other BCI models (both CNN and non-CNN based models) used in our model comparison. Section \ref{results} presents the results of both within-subject and cross-subject classification performance, as well as results of our feature explainability analysis. We discuss our findings in more detail in the Discussion.

\section{Materials and Methods}\label{methods}

\subsection{Data Description}\label{datasets}

BCIs are generally categorized into two types, depending on the EEG feature of interest \cite{Lotte2015}: event-related and oscillatory. \textit{Event-Related Potential} (ERP) BCIs are designed to detect a high amplitude and low frequency EEG response to a known, time-locked external stimulus. They are generally robust across subjects and contain well-stereotyped waveforms, enabling the time course of the ERP to be modeled through machine learning efficiently \cite{Fazel2012}. In contrast to ERP-based BCIs, which rely mainly on the detection of the ERP waveform from some external event or stimulus, \textit{Oscillatory} BCIs use the signal power of specific EEG frequency bands for external control and are generally asynchronous \cite{Pfurtscheller2001}. When oscillatory signals are time-locked to an external stimulus, they can be represented through event-related spectral perturbation (ERSP) analyses \cite{Makeig1993}. Oscillatory BCIs are more difficult to train, generally due to the lower signal-to-noise ratio (SNR) as well as greater variation across subjects \cite{Pfurtscheller2001}. A summary of the data used in this manuscript can be found in Table \ref{Data-Description}

\subsubsection{Dataset 1: P300 Event-Related Potential (P300)}
The P300 event-related potential is a stereotyped neural response to novel visual stimuli \cite{Polich2007}. It is commonly elicited with the visual oddball paradigm, where participants are shown repetitive ``non-target'' visual stimuli that are interspersed with infrequent ``target'' stimuli at a fixed presentation rate (for example, 1 Hz). Observed over the parietal cortex, the P300 waveform is a large positive deflection of electrical activity observed approximately 300 ms post stimulus onset, the strength of the observed deflection being inversely proportional to the frequency of the target stimuli. The P300 ERP is one of the strongest neural signatures observable by EEG, especially when targets are presented infrequently \cite{Polich2007}. When the image presentation rate increases to 2 Hz or more, it is commonly referred to as rapid serial visual presentation (RSVP), which has been used to develop BCIs for large image database triage \cite{Sajda2010, Marathe2016, Waytowich2016b}. 

The EEG data used here have been previously described in \cite{Marathe2016}; a brief description is given below. 18 participants volunteered for an RSVP BCI study. Participants were shown images of natural scenery at 2 Hz rate, with images either containing a vehicle or person (target), or with no vehicle or person present (non-target). Participants were instructed to press a button with their dominant hand when a target image was shown. The target/non-target ratio was $20\%/80\%$. Data from 3 participants were excluded from the analysis due to excessive artifacts and/or noise within the EEG data. Data from the remaining 15 participants (9 male and 14 right-handed) who ranged in age from 18 to 57 years (mean age 39.5 years) were further analyzed. EEG recordings were digitally sampled at 512 Hz from 64 scalp electrodes arranged in a 10-10 montage using a BioSemi Active Two system (Amsterdam, The Netherlands). Continuous EEG data were referenced offline to the average of the left and right earlobes, digitally bandpass filtered, using an FIR filter implemented in EEGLAB \cite{Delorme2004},  to 1-40 Hz and downsampled to 128 Hz. EEG trials of target and non-target conditions were extracted at $[0, 1]s$ post stimulus onset, and used for a two-class classification.

\begin{table}[!t]
    \small
	\centering
	\begin{adjustbox}{width=1\textwidth}
	\begin{tabular}{c |c c c c c c} 
		Paradigm & Feature Type & Bandpass Filter & $\#$ of Subjects & Trials per Subject & $\#$ of Classes & Class Imbalance?  \\ [0.5ex] 
		\hline
		P300 & ERP & 1-40Hz & 15 & $\sim$ 2000  & 2 & Yes, $\sim$ 5.6:1\\
		ERN & ERP & 1-40Hz & 26 &  340  & 2 & Yes, $\sim$ 3.4:1 \\
		MRCP & ERP/Oscillatory & 0.1-40Hz & 13  &  $\sim$ 1100 & 2 & No \\
		SMR  & Oscillatory & 4-40Hz & 9 & 288 & 4 & No \\
	\end{tabular}
 	\end{adjustbox}

	\caption{Summary of the data collections used in this study. Class imbalance, if present, is given as odds; i.e.: an odds of 2:1 means the class imbalance is $2/3$ of the data for class 1 to $1/3$ of the data for class 2. For the P300 and ERN datasets, the class imbalance is subject-dependent; therefore, the odds is given as the average class imbalance over all subjects.  }\label{Data-Description}
\end{table}

\subsubsection{Dataset 2: Feedback Error-Related Negativity (ERN)}

Error-Related Negativity potentials are perturbations of the EEG following an erroneous or unusual event in the subject's environment or task. They can be observed in a variety of tasks, including time interval production paradigms \cite{Miltner1997} and in forced-choice paradigms \cite{Gehring1993, Falkenstein1991}. Here we focus on the feedback error-related negativity (ERN), which is an amplitude perturbation of the EEG following the perception of an erroneous feedback produced by a BCI. The feedback ERN is characterized as a negative error component approximately 350ms, followed by a positive component approximately 500ms, after visual feedback (see Figure 7 of \cite{Margaux2012} for an illustration). The detection of the feedback ERN provides a mechanism to infer, and to possibly correct in real-time, the incorrect output of a BCI. This two-stage system has been proposed as a hybrid BCI in \cite{Zander2009, Millan2010} and has been shown to improve the performance of a P300 speller in online applications \cite{Spuler2012}.

The EEG data used here comes from  \cite{Margaux2012} and was used in the ``BCI Challenge'' hosted by Kaggle (https://www.kaggle.com/c/inria-bci-challenge); a brief description is given below. 26 healthy participants (16 for training, 10 for testing) participated in a P300 speller task, a system which uses a random sequence of flashing letters, arranged in a $6 \times 6$ grid, to elicit the P300 response \cite{Krusienski2008}. The goal of the challenge was to determine whether the feedback of the P300 speller was correct or incorrect. The EEG data were originally recorded at 600Hz using 56 passive Ag/AgCl EEG sensors (VSM-CTF compatible system) following the extended 10-20 system for electrode placement. Prior to our analysis, the EEG data were band-pass filtered, using an FIR filter implemented in EEGLAB \cite{Delorme2004}, to 1-40 Hz and down-sampled to 128Hz. EEG trials of correct and incorrect feedback were extracted at $[0, 1.25]s$ post feedback presentation and used as features for a two-class classification.

\subsubsection{Dataset 3: Movement-Related Cortical Potential (MRCP)}
Some neural activities contain both ERP as well as an oscillatory components. One particular example of this is the movement-related cortical potential (MRCP), which can be elicited by voluntary movements of the hands and feet and is observable through EEG along the central and midline electrodes, contralateral to the hand or foot movement \cite{Toro1994,Pfurtscheller1977, Pfurtscheller1999, Liao2014}. The MRCP components can be seen before movement onset (a slow 0-5Hz readiness potential \cite{Barrett1986, Yilmaz2015} and an early desynchronization in the 10-12Hz frequency band), at movement onset (a slow motor potential \cite{Deecke1969, Yilmaz2015}), and after movement onset (a late synchronization of 20-30Hz activity approximately 1 second after movement execution). The MRCP has been used previously to develop motor control BCIs for both healthy and physically disabled patients \cite{Leuthardt2006, YomTov2003, Karimi2017} 

The EEG data used here have been previously described in \cite{Gordon2015}; a brief description is given below. In this study, 13 subjects performed self-paced finger movements using the left index, left middle, right index, or right middle fingers. The data was recorded using a 256 channel BioSemi Active II system at 1024 Hz. Due to extensive signal noise present in the data, the EEG data were first processed with the PREP pipeline \cite{PREP}. The data were referenced to linked mastoids, bandpass filtered, using an FIR filter implemented in EEGLAB \cite{Delorme2004}, between 0.1 Hz and 40 Hz, and then downsampled to 128 Hz. We further downsampled the channel space to the standard 64 channel BioSemi montage. The index and middle finger blocks for each hand were combined for binary classification of movements originating from the left or right hand. EEG trials of left and right hand finger movements were extracted at $[-.5, 1]s$ around finger movement onset and used for a two-class classification.

\subsubsection{Dataset 4: Sensory Motor Rhythm (SMR)}
A common control signal for oscillatory-based BCI is the sensorimotor rhythm (SMR), wherein mu (8-12Hz) and beta (18-26Hz) bands desynchronize over the sensorimotor cortex contralateral to an actual or imagined movement. The SMR is very similar to the oscillatory component of the MRCP. Although SMR-based BCIs can facilitate nuanced, endogenous BCI control, they tend to be weak and highly variable across and within subjects, conventionally demanding user-training (neurofeedback) and long calibration times (20 minutes) in order to achieve reasonable performance \cite{Lotte2015}. 

The EEG data used here comes from BCI Competition IV Dataset 2A \cite{Tangermann2012} (called the SMR dataset for the remainder of the manuscript). The data consists of four classes of imagined movements of left and right hands, feet and tongue recorded from 9 subjects. The EEG data were originally recorded using 22 Ag/AgCl electrodes, sampled at 250 Hz and bandpass filtered between 0.5 and 100Hz. We resampled the timeseries to 128 Hz, and follow the same EEG pre-processing procedure as described in \cite{Schirrmeister2017b}, using software that was provided by the authors.  For both the training and test sets we epoched the data at [0.5, 2.5] seconds post cue onset (the same window which was used in \cite{Lotte2015, Sakhavi2015}). Note that we make predictions for only this time range on the test set. We perform a four-class classification using accuracy as the summary measure.

\subsection{Classification Methods}\label{classification-methods}

\subsubsection{EEGNet: Compact CNN Architecture} \label{eegnet-description}

\begin{figure*}[!t]
	\centering
	\includegraphics[scale=.4]{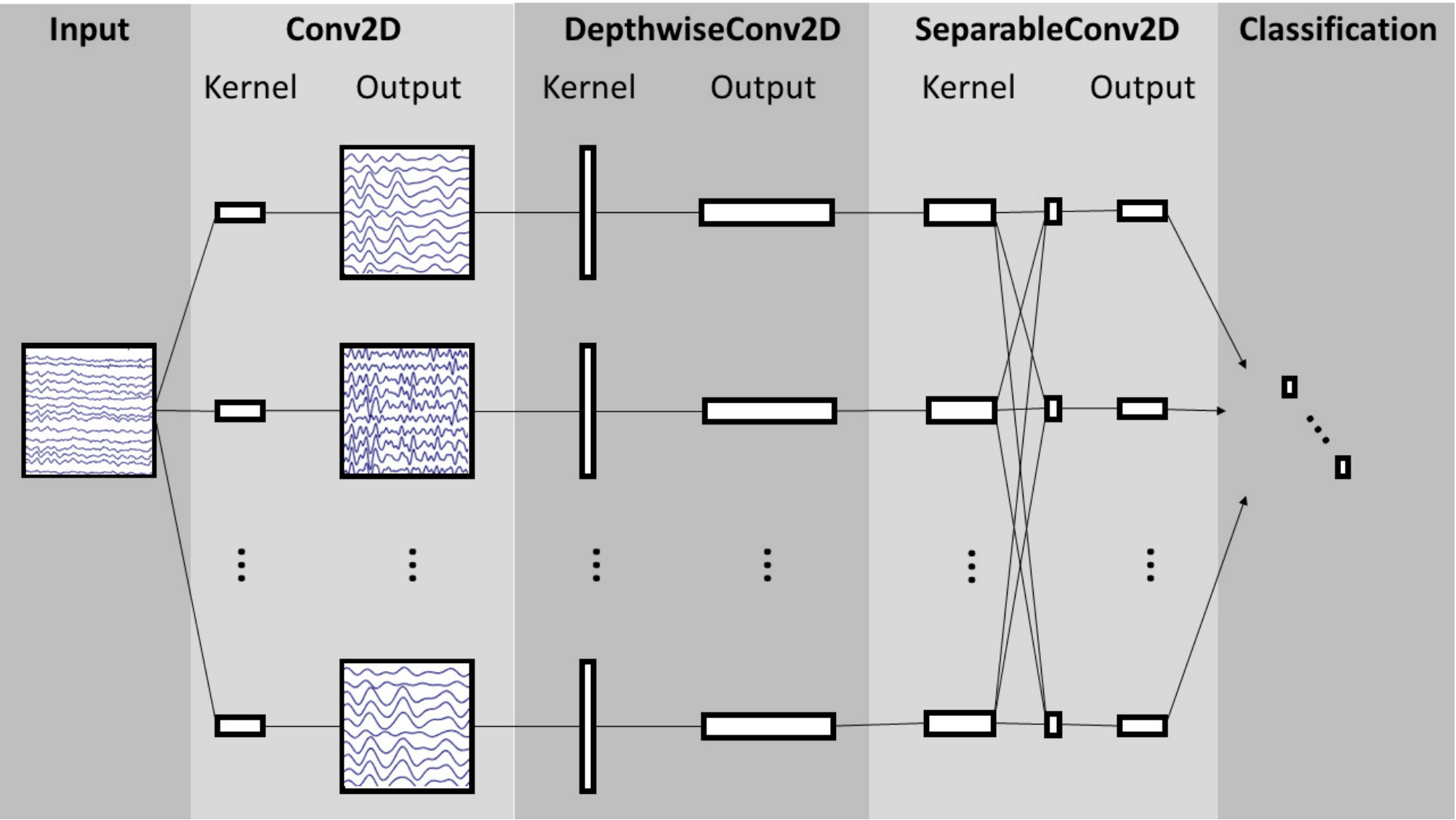}
	\caption{Overall visualization of the EEGNet architecture. Lines denote the convolutional kernel connectivity between inputs and outputs (called \textit{feature maps}) . The network starts with a temporal convolution (second column) to learn frequency filters, then uses a depthwise convolution (middle column), connected to each feature map individually, to learn frequency-specific spatial filters. The separable convolution (fourth column) is a combination of a depthwise convolution, which learns a temporal summary for each feature map individually, followed by a pointwise convolution, which learns how to optimally mix the feature maps together. Full details about the network architecture can be found in Table \ref{CNN-Model}.}
	\label{eegnet-architecture}
\end{figure*}

Here we introduce EEGNet, a compact CNN architecture for EEG-based BCIs that (1) can be applied across several different BCI paradigms, (2) can be trained with very limited data and (3) can produce neurophysiologically interpretable features. A visualization and full description of the EEGNet model can be found in Figure \ref{eegnet-architecture} and Table \ref{CNN-Model}, respectively, for EEG trials, collected at 128Hz sampling rate, having $C$ channels and $T$ time samples. We fit the model using the Adam optimizer, using default parameters as described in \cite{Kingma2014}, minimizing the categorical cross-entropy loss function. We run 500 training iterations (epochs) and perform validation stopping, saving the model weights which produced the lowest validation set loss. All models were trained on an NVIDIA Quadro M6000 GPU, with CUDA 9 and cuDNN v7, in Tensorflow \cite{Abadi2016b}, using the Keras API \cite{keras}. We omit the use of bias units in all convolutional layers. Note that, while all convolutions are one-dimensional, we use two-dimensional convolution functions for ease of software implementation. Our software implementation can be found at https://github.com/vlawhern/arl-eegmodels.

\begin{itemize}
\item In Block 1, we perform two convolutional steps in sequence. First, we fit $F_1$ 2D convolutional filters of size $(1, 64)$, with the filter length chosen to be half the sampling rate of the data (here, 128Hz), outputting $F_1$ feature maps containing the EEG signal at different band-pass frequencies. Setting the length of the temporal kernel at half the sampling rate allows for capturing frequency information at 2Hz and above. We then use a \textit{Depthwise Convolution} \cite{Chollet16a} of size $(C, 1)$ to learn a spatial filter. In CNN applications for computer vision the main benefit of a depthwise convolution is reducing the number of trainable parameters to fit, as these convolutions are not fully-connected to all previous feature maps (see Figure \ref{eegnet-architecture} for an illustration). Importantly, when used in EEG-specific applications, this operation provides a direct way to learn spatial filters for each temporal filter, thus enabling the efficient extraction of frequency-specific spatial filters (see the middle column of Figure \ref{eegnet-architecture}). A depth parameter $D$ controls the number of spatial filters to learn for each feature map ($D = 1$ is shown in Figure \ref{eegnet-architecture} for illustration purposes). This two-step convolutional sequence is inspired in part by the Filter-Bank Common Spatial Pattern (FBCSP) algorithm \cite{Ang2012} and is similar in nature to another decomposition technique, Bilinear Discriminant Component Analysis \cite{dyrholm2007}. We keep both convolutions linear as we found no significant gains in performance when using nonlinear activations. We apply Batch Normalization \cite{Ioffe15} along the feature map dimension before applying the exponential linear unit (ELU) nonlinearity \cite{Clevert15}. To help regularize or model, we use the Dropout technique \cite{Srivastava2014}. We set the dropout probability to $0.5$ for within-subject classification to help prevent over-fitting when training on small sample sizes, whereas we set the dropout probability to $0.25$ in cross-subject classification, as the training set sizes are much larger (see Section \ref{data-analysis} for more details on our within- and cross-subject analyses). We apply an average pooling layer of size (1, 4) to reduce the sampling rate of the signal to 32Hz. We also regularize each spatial filter by using a maximum norm constraint of 1 on its weights; $\left\| w \right\|^2 < 1$.

 \begin{table*}[t!]
 	\centering
 	\begin{adjustbox}{width=1\textwidth}
 	
 	\def\arraystretch{1.25}
 	\begin{tabular}{c|lllllllll}
 		\textbf{Block} & \textbf{Layer} & \textbf{\# filters} & \textbf{size}    & \textbf{\# params}  &  \textbf{Output} & \textbf{Activation} & \textbf{Options} \\ \hline
 		  
         1      & Input               	&                     &                       &                           & (C, T)                     &                     &               \\
                & Reshape             	&                     &                       &                           & (1, C, T)                  &                     &               \\
                & Conv2D              	& $F_1$               & (1, 64)               & $64 * F_1$                & ($F_1$, C, T)                  & Linear              	& mode = same          \\
                & BatchNorm           	&                     &                       & $2 * F_1$                 & ($F_1$, C, T)                  &                     &               \\
                & DepthwiseConv2D     	& D * $F_1$           & (C, 1)                & $C * D * F_1$             & (D * $F_1$, 1, T)                  & Linear              & mode = valid, depth = D, max norm = 1         \\
                & BatchNorm           	&                     &                       & $2 * D * F_1$             & (D * $F_1$, 1, T)                  &                     &               \\
                & Activation          	&                     &                       &                           & (D * $F_1$, 1, T)                  & ELU                 &               \\
				& AveragePool2D       	&                     & (1, 4)                &                           & (D * $F_1$, 1, T // 4)        &                     &               \\
                & Dropout*              &                     &                       &                           & (D * $F_1$, 1, T // 4)                  &                     &  $p = 0.25$ or $p = 0.5$            \\
         2      & SeparableConv2D     	& $F_2$               & (1, 16)               & $16 * D * F_1 + F_2 * (D * F_1)$    & ($F_2$, 1, T // 4)                  & Linear              & mode = same          \\
                & BatchNorm           	&                     &                       & $2 * F_2$                 & ($F_2$, 1, T // 4)                  &                     &               \\
                & Activation          	&                     &                       &                           & ($F_2$, 1, T // 4)                  & ELU                 &               \\
                & AveragePool2D       	&                     & (1, 8)                &                           & ($F_2$, 1, T // 32)             &                     &               \\
                & Dropout*              &                     &                       &                           & ($F_2$, 1, T // 32)             &                     &  $p = 0.25$ or $p = 0.5$              \\
		        & Flatten             	&                     &                       &                           & ($F_2$ * (T // 32))        &                     &               \\
 		 Classifier     & Dense               	& N * ($F_2$ * T // 32) &                    &                          & N                          & Softmax             &  max norm = 0.25            
 	\end{tabular}
 	\end{adjustbox}
 	\vspace{3mm}
	 \caption{EEGNet architecture, where $C = $ number of channels,  $T = $ number of time points, $F_1 = $ number of temporal filters, $D = $ depth multiplier (number of spatial filters), $F_2 = $ number of pointwise filters, and $N = $ number of classes, respectively. For the Dropout layer, we use $p = 0.5$ for within-subject classification and $p = 0.25$ for cross-subject classification (see Section 2.1.1 for more details)} \label{CNN-Model}
 \end{table*}

\item In Block 2, we use a \textit{Separable Convolution}, which is a Depthwise Convolution (here, of size $(1, 16)$, representing 500ms of EEG activity at 32Hz) followed by $F_2$ $(1, 1)$ Pointwise Convolutions \cite{Chollet16a}. The main benefits of separable convolutions are (1) reducing the number of parameters to fit and (2) explicitly decoupling the relationship within and across feature maps by first learning a kernel summarizing each feature map individually, then optimally merging the outputs afterwards. When used for EEG-specific applications this operation separates learning how to summarize individual feature maps in time (the depthwise convolution) with how to optimally combine the feature maps (the pointwise convolution). This operation is also particularly useful for EEG signals as different feature maps may represent data at different time-scales of information. In our case we first learn a 500 ms ``summary'' of each feature map, then combine the outputs afterwards.  An Average Pooling layer of size $(1, 8)$ is used for dimension reduction.

\item In the classification block, the features are passed directly to a softmax classification with $N$ units, $N$ being the number of classes in the data. We omit the use of a dense layer for feature aggregation prior to the softmax classification layer to reduce the number of free parameters in the model, inspired by the work in \cite{Springenberg2014}.

\end{itemize}

We investigate several different configurations of the EEGNet architecture by varying the number of filters, $F_1$, and the number of spatial filters per temporal filter, $D$ to learn. We set $F_2 = D * F_1$ (the number of temporal filters along with their associated spatial filters from Block 1) for the duration of the manuscript, although in principle $F_2$ can take any value; $F_2 < D * F_1$ denotes a compressed representation, learning fewer feature maps than inputs, whereas $F_2 > D * F_1$ denotes an overcomplete representation, learning more feature maps than inputs. We use the notation EEGNet-$F_1$,D to denote the number of temporal and spatial filters to learn; i.e.: EEGNet-4,2 denotes learning 4 temporal filters and 2 spatial filters per temporal filter.

\subsubsection{Comparison with existing CNN Approaches} 

We compare the performance of EEGNet against the DeepConvNet and ShallowConvNet models proposed by \cite{Schirrmeister2017b}; full table descriptions of both models can be found in the Appendix. We implemented these models in Tensorflow and Keras, following the descriptions found in the paper. As their architectures were originally designed for 250Hz EEG signals (as opposed to 128Hz signals used here) we divided the lengths of temporal kernels and pooling layers in their architectures by 2 to correspond approximately to the sampling rate used in our models. We train these models in the same way we train the EEGNet model (see Section \ref{eegnet-description}). 

The DeepConvNet architecture consists of five convolutional layers with a softmax layer for classification (see Figure 1 of \cite{Schirrmeister2017b}). The ShallowConvNet architecture consists of two convolutional layers (temporal, then spatial), a squaring nonlinearity ($f(x) = x^2$), an average pooling layer and a log nonlinearity ($f(x)=\log(x)$). We would like to emphasize that the ShallowConvNet architecture was designed specifically for oscillatory signal classification (by extracting features related to log band-power); thus, it may not work well on ERP-based classification tasks. However, the DeepConvNet architecture was designed to be a general-purpose architecture that is not restricted to specific feature types \cite{Schirrmeister2017b}, and thus it serves as a more valid comparison to EEGNet. Table \ref{model-size-comparison} shows the number of trainable parameters per model across all CNN models.

\begin{table}[!t]
	\centering
	\begin{tabular}{ l |c c c c c} 
		& Trial Length (sec)  & DeepConvNet & ShallowConvNet &  EEGNet-4,2 & EEGNet-8,2 \\ [0.5ex] 
		\hline
		P300         & 1 & 174,127 & 104,002  &  \textbf{1,066}   & 2,258 \\
		ERN & 1.25 & 169,927 & 91,602 & \textbf{1,082}   & 2,290  \\
		MRCP   & 1.5 & 175,727 & 104,722 & \textbf{1,098}   & 2,322 \\
		SMR*  & 2 & 152,219  & 40,644 & \textbf{796} & 1,716  \\
	\end{tabular}
	\caption{Number of trainable parameters per model and per dataset for all CNN-based models. We see that the EEGNet models are up to two orders of magnitude smaller than both DeepConvNet and ShallowConvNet across all datasets. Note that we use a temporal kernel length of 32 samples for the SMR dataset as the data were high-passed at 4Hz.}
	\label{model-size-comparison}
\end{table}

\subsubsection{Comparison with Traditional  Approaches}
We also compare the performance of EEGNet to that of the best performing traditional approach for each individual paradigm.  For all ERP-based data analyses (P300, ERN, MRCP) the traditional approach is the approach which won the Kaggle BCI Competition (code and documentation at http://github.com/alexandrebarachant/bci-challenge-ner-2015), which uses a combination of xDAWN Spatial Filtering \cite{Rivet2009}, Riemannian Geometry \cite{barachant_multiclass_2012, barachant_plug&play_2014}, channel subset selection and $L_1$ feature regularization (referred to as xDAWN + RG for the remainder of the manuscript). Here we provide a summary of the approach, which is done in five steps:

\begin{itemize}
    \item[1.] Train two set of 5 xDAWN spatial filters, one set for each class of a binary classification task, using the ERP template concatenation method as described in  \cite{barachant_plug&play_2014, CongedoBA13}.
    \item[2.] Perform EEG electrode selection through backward elimination \cite{Barachant2011} to keep only the most relevant 35 channels.
    \item[3.] Project the covariance matrices onto the tangent space using the log-euclidean metric \cite{barachant_multiclass_2012, barachant_classification_2013}.
    \item[4.] Perform feature normalization using an $L_1$ ratio of 0.5, signifying an equal weight for $L_1$ and $L_2$ penalties. An $L_1$ penalty encourages the sum of the absolute values of the parameters to be small, whereas an $L_2$ penalty encourages the sum of the squares of the parameters to be small (a theoretical overview of these penalties can be found in \cite{Ng2004}).
    \item[5.] Perform classification using an Elastic Net regression.
\end{itemize}

\noindent We use the same xDAWN+RG model parameters across all comparisons (P300, ERN, MRCP) with the exception of the initial number of EEG channels to use, which was set to 56 for ERN and 64 for P300 and MRCP. While the original solution used an ensemble of bagged classifiers, for this analysis we only compared a single model with this approach to a single EEGNet model on identical training and test sets, as we expect any gains from ensemble learning to benefit both approaches equally. The original solution also used a set of ``meta features'' that were specific to that data collection. As the goal of this work is to investigate a general-purpose CNN model for EEG-based BCIs, we omitted the use of these features as they are specific to that particular data collection. 

For oscillatory-based classification of SMR, the traditional approach is our own implementation of the One-Versus-Rest (OVR) filter-bank common spatial pattern (FBCSP) algorithm as described in \cite{Ang2012}. Here we provide a brief summary of our approach:

\begin{itemize}
    \item[1.] Bandpass filter the EEG signal into 9 non-overlapping filter banks in 4Hz steps, starting at 4Hz: 4-8Hz, 8-12Hz, ..., 36-40Hz.
    \item[2.] As the classification problem is multi-class, we use OVR classification, which requires that we train a classifier for all pairs of OVR combinations, which there are 4 here (class 1 vs all others, class 2 vs all others, etc). We train 2 CSP filter pairs (4 filters total) for each filter bank on the training data using the auto-covariance shrinkage method by \cite{Ledoit2004}. This will give a total of 36 features (9 filter banks $\times$ 4 CSP filters) for each trial and each OVR combination.
    \item[3.] Train an elastic-net logistic regression classifier \cite{Zou2005} for each OVR combination. We set the elastic net penalty $\alpha = 0.95$.
    \item[4.] Find the optimal $\lambda$ value for the elastic-net logistic regression that maximizes the validation set accuracy by evaluating the trained classifiers on a held-out validation set. The multi-class label for each trial is the classifier that produces the highest probability among the 4 OVR classifiers.
    \item[5.] Apply the trained classifiers to the test set, using the $\lambda$ values obtained in Step 4.
\end{itemize}

\noindent Note that this approach differs slightly from the original technique as proposed in \cite{Ang2012}, where they use a Naive Bayes Parzen Window classifier. We opted to use an elastic net logistic regression for ease of implementation, and the fact that it has been used in existing software implementations of FBCSP (for example, in BCILAB \cite{Kothe2013}). 
  
\subsection{Data Analysis} \label{data-analysis}

Classification results are reported for two sets of analyses: within-subject and cross-subject. Within-subject classification uses a portion of the subjects data to train a model specifically for that subject, although cross-subject classification uses the data from other subjects to train a subject-agnostic model. While within-subject models tend to perform better than cross-subject models on a variety of tasks, there is ongoing research investigating techniques to minimize (or possibly eliminate) the need for subject-specific information to train robust systems \cite{Lotte2015, Waytowich2016b}. 

For within-subject, we use four-fold blockwise cross-validation, where two of the four blocks are chosen to be the training set, one block as the validation set, and the final block as testing. We perform statistical testing using a repeated-measures Analysis of Variance (ANOVA), modeling classification results (AUC for P300/MRCP/ERN and Classification Accuracy for SMR) as the response variable with subject number and classifier type as factors. For cross-subject analysis in P300 and MRCP we choose, at random, 4 subjects for the validation set, one subject for the test set, and all remaining subjects for the training set (see Table \ref{Data-Description} for number of subjects per dataset). This process was repeated 30 times, producing 30 different folds. We follow the same procedure for the ERN dataset, except we use the 10 test subjects from the original Kaggle Competition as the test set for each fold. We perform statistical testing using a one-way Analysis of Variance, using classifier type as the factor. For the SMR dataset, we partitioned the data as follows: For each subject, select the training data from 5 other subjects at random to be the training set and the training data from the remaining 3 subjects to be the validation set. The test set remains the same as the original test set for the competition. Note that this enforces a fully cross-subject classification analysis as we never use the test subjects' training data. This process is repeated 10 times for each subject, creating 90 different folds.  The mean and standard error of classification performance were calculated over the 90 folds. We perform statistical testing for this analysis using the same testing procedure as the within-subject analysis.   

When training both the within-subject and cross-subject models, we apply a class-weight to the loss function whenever the data is imbalanced (unequal number of trials for each class). The class-weight we apply is the inverse of the proportion in the training data, with the majority class set to 1. For example, in the P300 dataset, there is a 5.6:1 odds between non-targets and targets (Table \ref{Data-Description}) . In this case the class-weight for non-targets was set to 1, while the class-weight for targets was set to 6 (when the odds are a fraction, we take the next highest integer). This procedure was applied to the P300 and ERN datasets only, as these were the only datasets where significant class imbalance was present.

Note that for the SMR analysis, we set the temporal kernel length to be 32 samples long (as opposed to 64 samples long as given in Table \ref{CNN-Model}) since the data were high-passed at 4Hz.

\subsection{EEGNet Feature Explainability}
The development of methods for enabling feature explainability from deep neural networks has become an active research area over the past few years, and has been proposed as an essential component of a robust model validation procedure, to ensure that the classification performance is being driven by relevant features as opposed to noise or artifacts in the data \cite{ baehrens2010explain, Zeiler2014, SimonyanZ14a, Nguyen2014, Ribeiro2016, ShrikumarGK17, ancona2018, Montavon2018}. We present three different approaches for understanding the features derived by EEGNet: 

\begin{itemize} 
	\item[1.] \textbf{Summarizing averaged outputs of hidden unit activations:} This approach focuses on summarizing the activations of hidden units at layers specified by the user. In this work we choose to summarize the hidden unit activations representing the data after the depthwise convolution (the spatial filter operation in EEGNet). Because the spatial filters are tied directly to a particular temporal filter, they provide additional insights into the spatial localization of narrow-band frequency activity.  Here we summarize the spatially-filtered data by calculating the difference in averaged time-frequency representations between classes, using Morlet wavelets \cite{torrence1998practical}.  
		
	\item[2.] \textbf{Visualizing the convolutional kernel weights:} This approach focuses on directly visualizing and interpreting the convolutional kernel weights from the model. Generally speaking, interpreting the convolutional kernel weights is very difficult due to the cross-filter-map connectivity between any two layers. However, because EEGNet limits the connectivity of the convolutional layers (using depthwise and separable convolutions), it is possible to interpret the temporal convolution as narrow-band frequency filters and the depthwise convolution as frequency-specific spatial filters.

	\item[3.] \textbf{Calculating single-trial feature \textit{relevance} on the classification decision:} This approach focuses on calculating, on a single-trial basis, the \textit{relevance} of individual features on the resulting classification decision. Positive values of relevance denote evidence supporting the outcome, while negative values of relevance denote evidence against the outcome. In our analysis we used DeepLIFT with the Rescale rule \cite{ShrikumarGK17}, as implemented in \cite{ancona2018}, to calculate single-trial EEG feature relevance. DeepLIFT is a gradient-based relevance attribution method that calculates relevance values per feature relative to a ``reference'' input (here, an input of zeros, as is suggested in \cite{ShrikumarGK17}), and is a technique similar to Layerwise Relevance Propagation (LRP) which has been used previously for EEG analysis \cite{Sturm2016} (a summary of gradient-based relevance attribution methods can be found in \cite{ancona2018}). This analysis can be used to elucidate feature relevance from high-confidence versus low-confidence predictions, and can be used to confirm that the relevant features learned are interpretable, as opposed to noise or artifact features.
\end{itemize}

\section{Results}\label{results}
 
\subsection{Within-Subject Classification}

\begin{figure*}[!t]
	\centering
	\includegraphics[scale=.4]{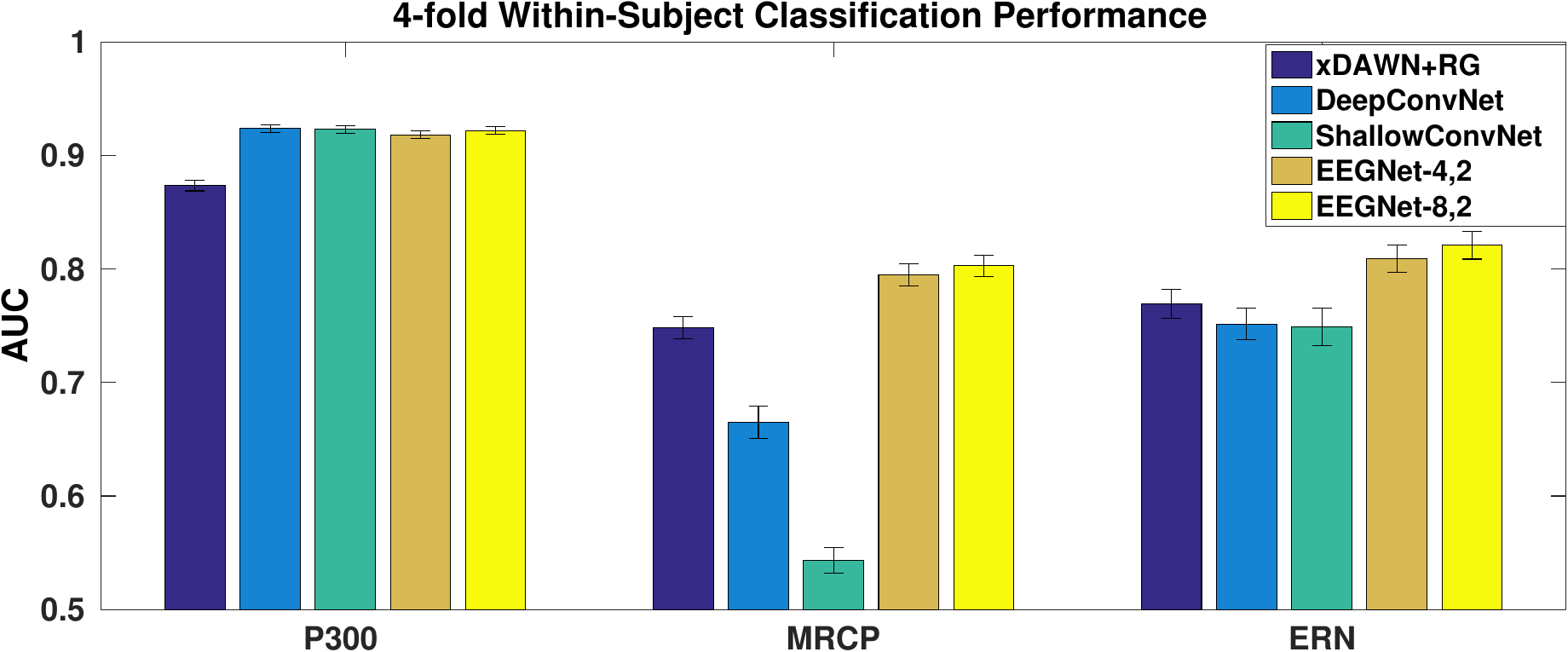}
	\caption{4-fold within-subject classification performance for the P300, ERN and MRCP datasets for each model, averaged over all folds and all subjects. Error bars denote 2 standard errors of the mean. We see that, while there is minimal difference between all the CNN models for the P300 dataset, there are significant differences in the MRCP dataset, with both EEGNet models outperforming all other models. For the ERN dataset we also see both EEGNet models performing better than all others $(p < 0.05)$.}
	\label{within-subject}
\end{figure*}

We compare the performance of both the CNN-based reference algorithms (DeepConvNet and ShallowConvNet) and the traditional approach (xDAWN+RG for P300/MRCP/ERN and FBCSP for SMR) with EEGNet-4,2 and EEGNet-8,2. Within-subject four-fold cross-validation results across all algorithms for P300, MRCP and ERN datasets are shown in Figure \ref{within-subject}. We observed, across all paradigms, that there was no statistically significant difference between EEGNet-4,2 and EEGNet-8,2 $(p > 0.05)$, indicating that the increase in model complexity did not statistically improve classification performance. For the P300 dataset, all CNN-based models significantly outperform xDAWN+RG $(p < 0.05)$  while not performing significantly different amongst themselves. For the ERN dataset, EEGNet-8,2 outperforms DeepConvNet, ShallowConvNet and xDAWN+RG $(p < 0.05)$, while EEGNet-4,2 outperforms DeepConvNet and ShallowConvNet $(p < 0.05)$. The biggest difference observed among all the approaches is in the MRCP dataset, where both EEGNet models statistically outperform all others by a significant margin (DeepConvNet, ShallowConvNet and xDAWN+RG, $p < 0.05$ for each comparison).

\begin{figure*}[!t]
	\centering
	\includegraphics[scale=.35]{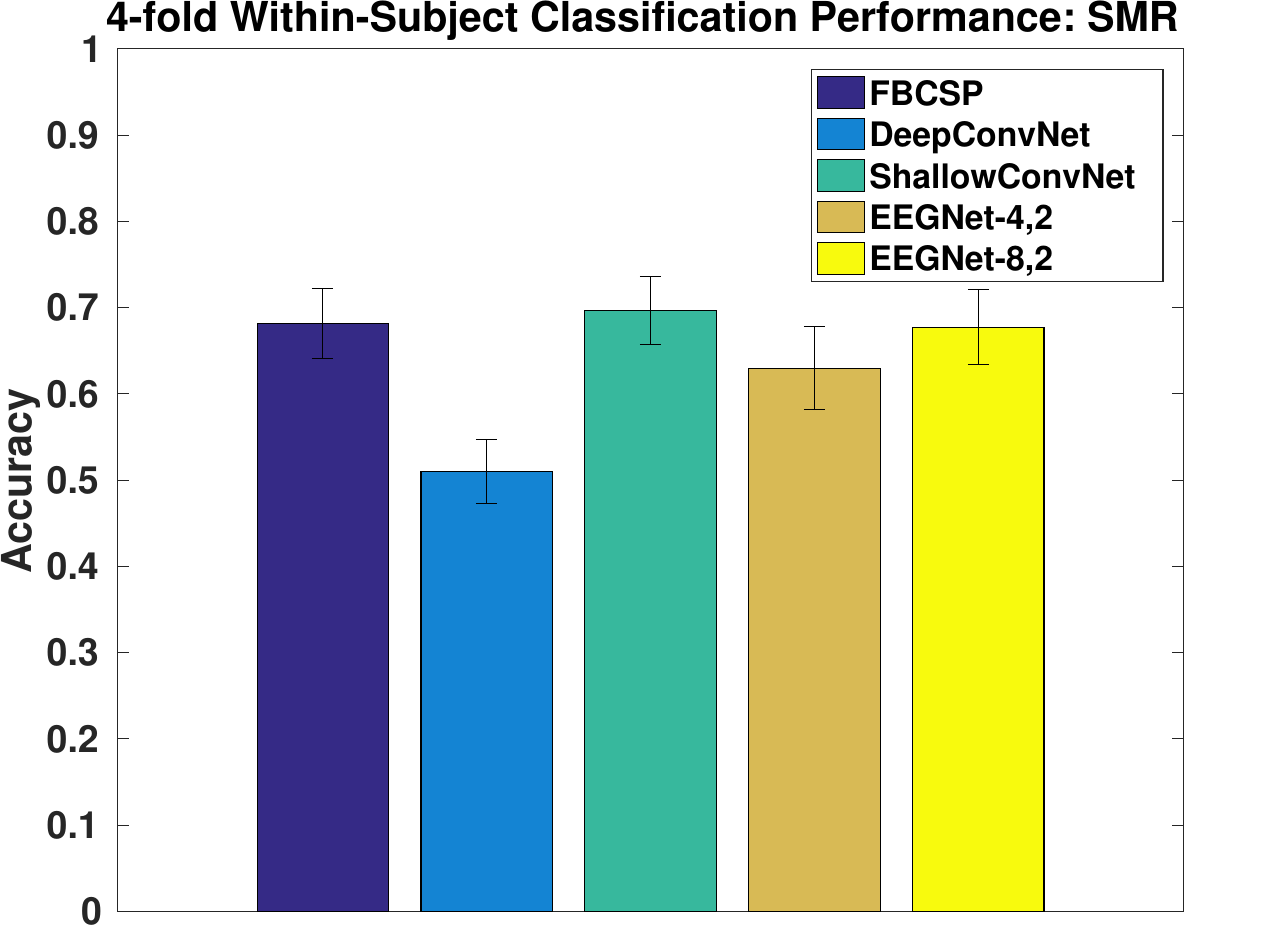}
	\caption{4-fold within-subject classification performance for the SMR dataset for each model, averaged over all folds and all subjects. Error bars denote 2 standard errors of the mean. Here we see DeepConvNet statistically performed worse than all other models $(p < 0.05)$. ShallowConvNet and EEGNet-8,2 performed similarly to that of FBCSP. }
	\label{within-subject-smr}
\end{figure*}

Four-fold cross-validation results for the SMR dataset are shown in Figure \ref{within-subject-smr}. Here we see the performances of ShallowConvNet and FBCSP are very similar, replicating previous results as reported in \cite{Schirrmeister2017b}, while DeepConvNet performance is significantly lower. We also see that EEGNet-8,2 performance is similar to FBCSP as well.

\subsection{Cross-Subject Classification}

\begin{figure*}[!b]
	\centering
	\includegraphics[scale=.4]{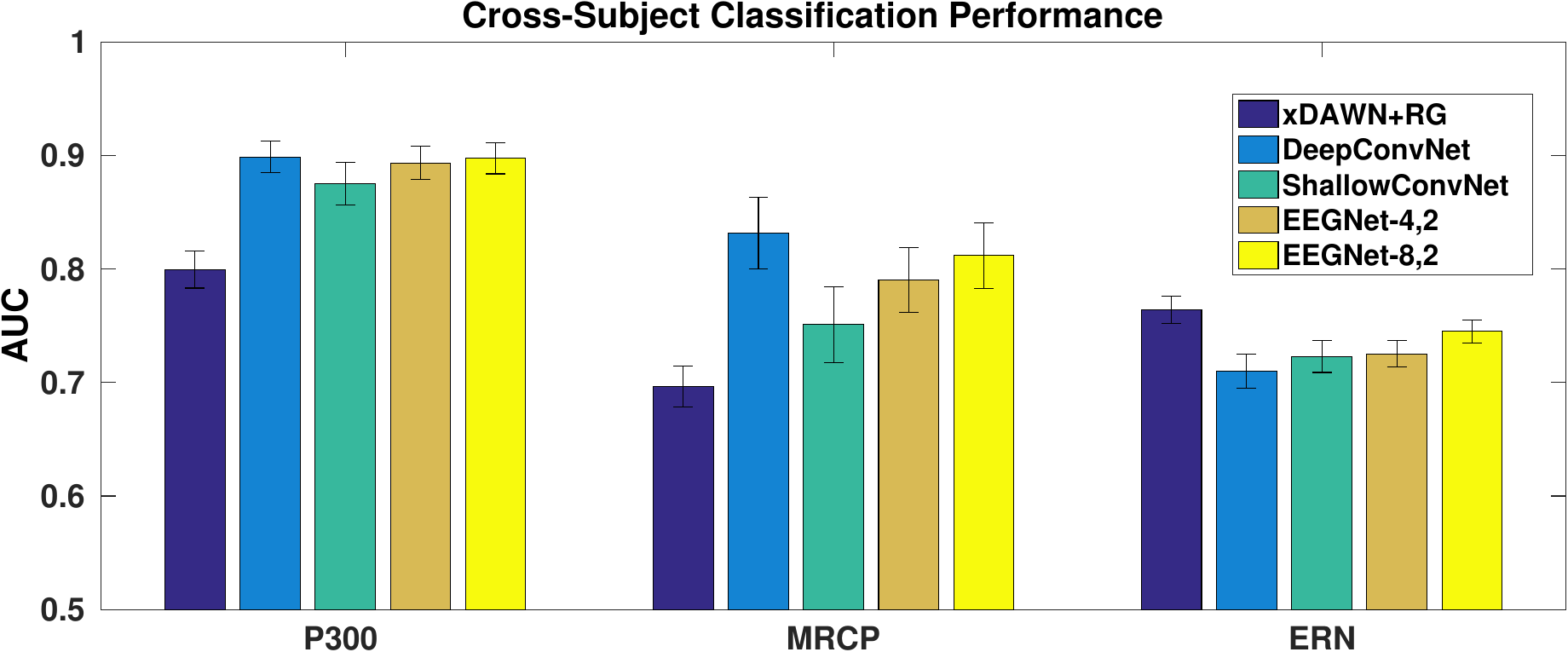}
	\caption{Cross-Subject classification performance for the P300, ERN and MRCP datasets for each model, averaged for 30 folds. Error bars denote 2 standard errors of the mean. For the P300 and MRCP datasets there is minimal difference between the DeepConvNet and the EEGNet models, with both models outperforming ShallowConvNet. For the ERN dataset the reference algorithm (xDAWN + RG) significantly outperforms all other models. }
	\label{cross-subject}
\end{figure*}

\begin{figure*}[!t]
	\centering
	\includegraphics[scale=.35]{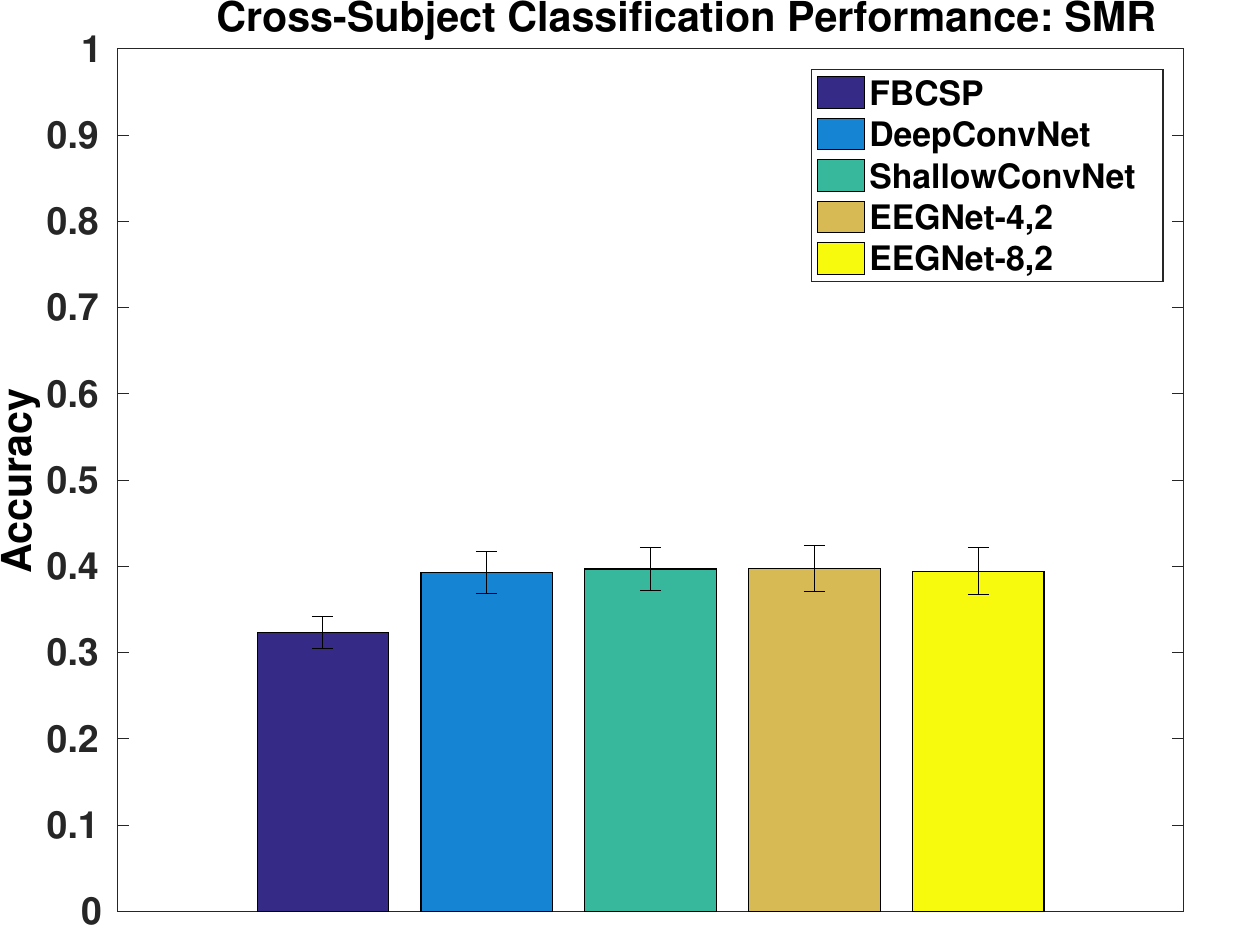}
	\caption{Cross-Subject classification performance for the SMR for each model, averaged over all folds and all subjects. Error bars denote 2 standard errors of the mean. We see that all CNN-based models perform similarly, while slightly outperforming FBCSP.}
	\label{cross-subject-smr}
\end{figure*}

Cross-subject classification results across all algorithms for P300, MRCP and ERN datasets are shown in Figure \ref{cross-subject}. Similar to the within-subject analysis, we observed no statistical difference between EEGNet-4,2 and EEGNet-8,2 across all datasets $(p > 0.05)$. For the P300 dataset, all CNN-based models significantly outperform xDAWN+RG $(p < 0.05)$  while not performing significantly different amongst themselves. For the MRCP dataset EEGNet-8,2 and DeepConvNet significantly outperform ShallowConvNet ($p < 0.05$). We also see that both DeepConvNet and ShallowConvNet performance is better when compared to its within-subject performance for the MRCP dataset. For the ERN dataset, xDAWN + RG outperforms all CNN models $(p < 0.05)$. Cross-subject classification results for the SMR dataset are shown in Figure \ref{cross-subject-smr}, where we found no significant difference in performance across all CNN-based models ($p > 0.05$).

\subsection{EEGNet Feature Characterization}

We illustrate three different approaches to characterize the features learned by EEGNet: (1) Summarizing averaged outputs of hidden unit activations, (2) visualizing convolutional kernel weights, and (3) calculating single-trial feature relevances on classification decision. We illustrate Approach 1 on the P300 dataset for a cross-subject trained EEGNet-4,1 model. We chose to analyze the filters from the P300 dataset due to the fact that multiple neurophysiological events occur simultaneously: participants were told to press a button with their dominant hand whenever a target image appeared on the screen. Because of this, target trials contain both the P300 event-related potential as well as the alpha/beta desynchronizations in contralateral motor cortex due to button presses. Here we were interested in whether or not the EEGNet architecture was capable of separating out these confounding events. We were also interested in quantifying the classification performance of the architecture whenever specific filters were removed from the model.

\begin{figure*}[!t]
	\centering
	\includegraphics[scale=0.775]{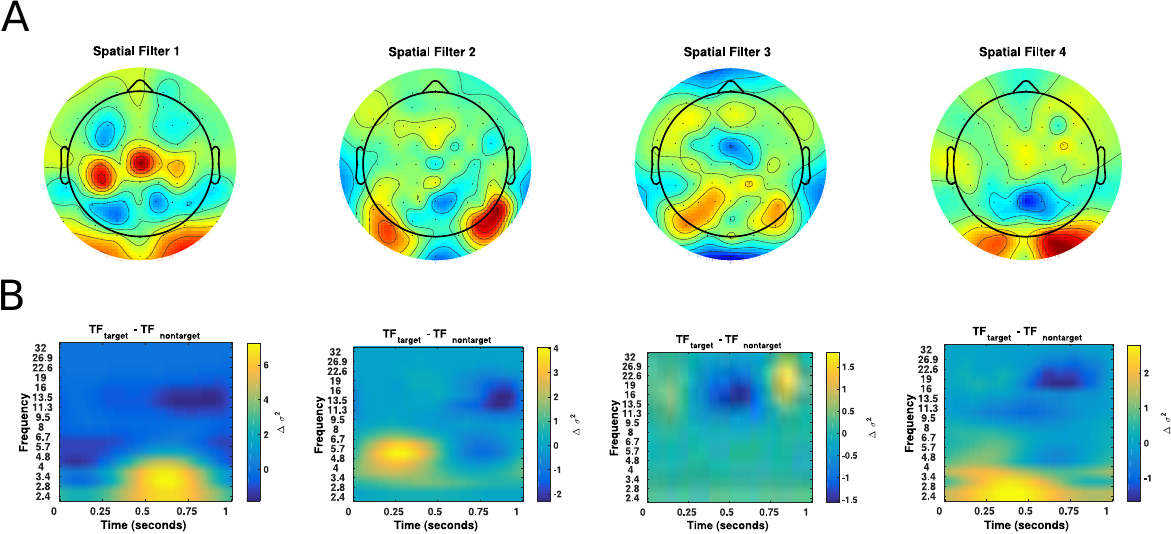}
	\caption{Visualization of the features derived from an EEGNet-4,1 model configuration for one particular cross-subject fold in the P300 dataset. (A) Spatial topoplots for each spatial filter. (B) The mean wavelet time-frequency difference between target and non-target trials for each individual filter. }
	\label{p300-visualization}
\end{figure*}

\begin{table}[!b]
	\scriptsize
	\centering
	\begin{tabular}{|c  c|} 
		\hline
		Filters Removed  & Test Set AUC \\ [0.5ex] 
		\hline
		(1) 		& 0.8866 \\
		(2) 		& \textbf{0.9076}\\
		(3) 		& 0.8910 \\
		(4) 		& 0.8747 \\
		\hline
		(1, 2)  	& 0.8875 \\
		(1, 3)  	& 0.8593 \\
		(1, 4)  	& 0.8325 \\
		(2, 3)  	& \textbf{0.8923} \\
		(2, 4)  	& 0.8721 \\
		(3, 4)  	& 0.8206 \\
		\hline
		(1, 2, 3) 	& \textbf{0.8637} \\
		(1, 2, 4) 	& 0.8202 \\
		(1, 3, 4) 	& 0.7108 \\
		(2, 3, 4) 	& 0.7970 \\
		\hline
		None        & \textbf{0.9054} \\	
		\hline
	\end{tabular}
	\caption{Performance of a cross-subject trained EEGNet-4,1 model when removing certain filters from the model, then using the model to predict the test set for one randomly chosen fold of the P300 dataset. AUC values in bold denote the best performing model when removing 1, 2 or 3 filters at a time. As the number of filters removed increases, we see decreases in classification performance, although the magnitude of the decrease depends on which filters are removed. }
	\label{p300-ablation}
\end{table}

Figure \ref{p300-visualization} shows the spatial topographies of the four filters along with an average wavelet time-frequency difference, calculated using Morlet wavelets \cite{torrence1998practical}, between all target trials and all non-target trials. Here we see four distinct filters appear. The time-frequency analysis of Filter 1 shows an increase in low-frequency power approximately 500ms after image presentation, followed by desynchronizations in alpha frequency. As nearly all subjects in the P300 dataset are right-handed, we also see significant activity along the left motor cortex. Time-frequency analysis of Filter 2 appears to show a significant theta-beta relationship; while increases in theta activity have been previously noted in the P300 literature in response to targets \cite{Mazaheri2005}, a relationship between theta and beta has not previously been noted. The time-frequency difference for Filter 4 appears to correspond with the P300, with an increase low-frequency power approximately 350ms after image presentation.

\begin{figure*}[!t]
	\centering
	\includegraphics[scale=1]{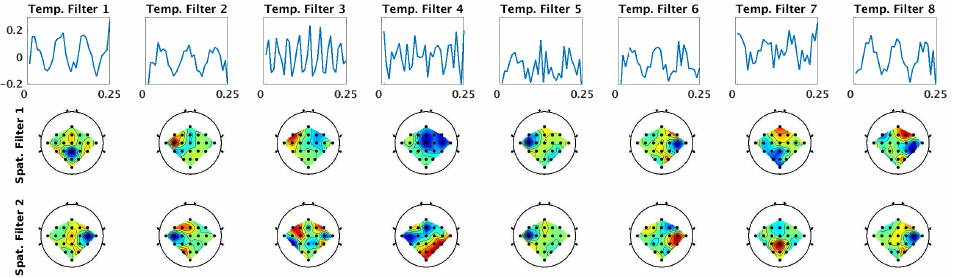}
	\caption{Visualization of the features derived from a within-subject trained EEGNet-8,2 model for Subject 3 of the SMR dataset. Each of the 8 columns shows the learned temporal kernel for a 0.25 second window (top) with its two associated spatial filters (bottom two). We see that, while many of the temporal filters are isolating slower-wave activity, the network identifies a higher-frequency filter at approximately 32Hz (Temp. Filter 3, which shows 8 cycles in a 0.25 s window). }
	\label{SMR-filter-vis}
\end{figure*}

We also conducted a feature ablation study, where we iteratively removed a set of filters (by replacing the filters with zeros) and re-applied the model to predict trials in the test set. We do this for all combinations of the four filters. Classification results for this ablation study are shown in Table \ref{p300-ablation}. We see that test set performance is minimally impacted by the removal of any single filter, with the largest decrease occurring when removing Filter 4. As expected, when removing pairs of filters the decrease in performance is more pronounced, with the largest decrease observed when removing Filters 3 and 4. Removing Filters 2 and 3 results in practically no change in classification performance when compared to the full model, suggesting that the most important features in this task are being captured by Filters 1 and 4. This finding is further reinforced when looking at classification performance when three filters are removed; a model that contains only Filter 4 (0.8637 AUC) performs fairly well when compared to models that contain only Filter 2 (0.7108 AUC) or Filter 1 (0.7970 AUC).

\begin{figure*}[!t]
	\centering
	\includegraphics[scale=0.85]{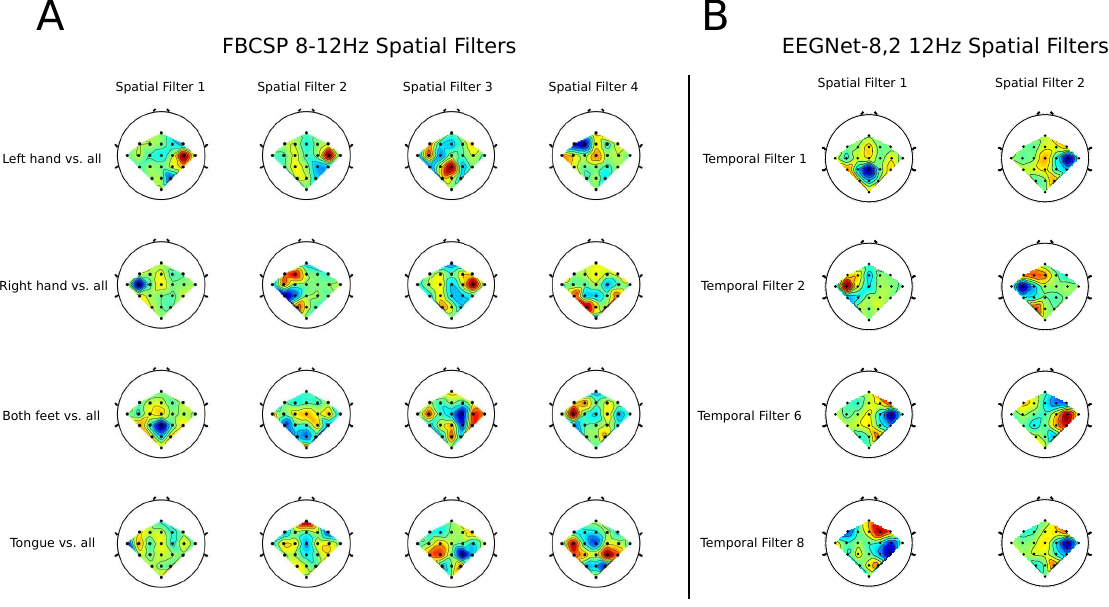}
	\caption{Comparison of the 4 spatial filters learned by FBCSP in the 8-12Hz filter bank for each OVR class combination (A) with the spatial filters learned by EEGNet-8,2 (B) for 4 temporal filters that capture 12Hz frequency activity for Subject 3 of the SMR dataset (Temporal Filters 1, 2, 6 and 8, see Figure \ref{SMR-filter-vis}). We see that similar filters appear across both FBCSP and EEGNet-8,2. }
	\label{SMR-spatialFilters}
\end{figure*}

 Figure \ref{SMR-filter-vis} shows the filters learned for the EEGNet-8,2 model for a within-subject classification of Subject 3 for the SMR dataset. Each column of this figure denotes the learned temporal kernel (top row) with its two associated spatial filters (bottom two rows).  Note that we are learning temporal filters of length 32 samples, which correspond to 0.25 seconds in time; hence, we estimate the frequency for each temporal filter as four times the number of observed cycles. Here we see that EEGNet-8,2 learns both slow-frequency activity at approximately 12Hz (Filters 1, 2, 6 and 8, which show three cycles in a 0.25 s window) and high-frequency activity at approximately 32Hz (Filter 3, which show 8 cycles). Figure \ref{SMR-spatialFilters} compares the spatial filters associated with 8-12Hz frequency band learned by EEGNet-8,2 with the spatial filters learned by FBCSP in the 8-12Hz filter-bank for each of the four OVR combinations. For ease of description we will use the notation X-Y to denote the row-column filter. Here we see many of the filters are strongly positively correlated across models (i.e.: the 1-1 filter of EEGNet-8,2 with the 3-1 filter for FBCSP ($\rho = 0.93$) and the 2-1 filter of EEGNet-8,2 with the 3-4 filter of FBCSP ($\rho = 0.83$)), while some are strongly negatively correlated (the 3-1 filter of EEGNet-8,2 with the 1-1 filter of FBCSP ($\rho = -0.93$)), indicating a similar filter up to a sign ambiguity.  This suggests that EEGNet, through the use of depthwise convolutions, is capable of learning band-specific spatial filters in a similar manner as FBCSP. 

\begin{figure*}[!t]
	\centering
	\includegraphics[scale=1.6]{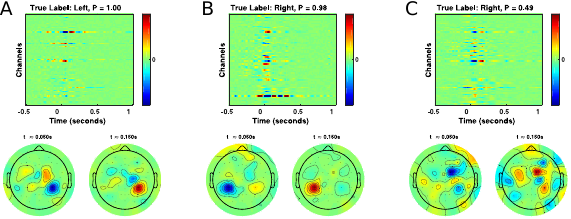}
	\caption{(Top row) Single-trial EEG feature relevance for a cross-subject trained EEGNet-8,2 model, using DeepLIFT, for three different test trials of the MRCP dataset: (A) a high-confidence, correct prediction of left finger movement, (B) a high-confidence, correct prediction of right finger movement and (C) a low-confidence, incorrect prediction of left finger movement. Titles include the true class label and the predicted probability of that label.  (Bottom row) Spatial topoplots of the relevances at two time points: approximately 50 ms and 150 ms after button press. As expected, the high-confidence trials show the correct relevances corresponding to contralateral motor cortex for left (A) and right (B) button presses, respectively. For the low-confidence trial we see the relevances are more mixed and broadly distributed, without a clear spatial localization to motor cortices. }
	\label{MRCP-DeepLIFT}
\end{figure*}

Figure \ref{MRCP-DeepLIFT} shows the single-trial feature relevances for EEGNet-8,2, calculated using DeepLIFT, for three three different test trials for one cross-subject fold of the MRCP dataset. Here we see that the high-confidence predictions (Figure \ref{MRCP-DeepLIFT}A and Figure \ref{MRCP-DeepLIFT}B, for left and right finger movement, respectively) both correctly show the contralateral motor cortex relevance as expected, whereas for a low-confidence prediction (Figure \ref{MRCP-DeepLIFT}C), the feature relevance is more broadly distributed, both in time and in space on the scalp. 

\begin{figure*}[!t]
	\centering
	\includegraphics[scale=0.45]{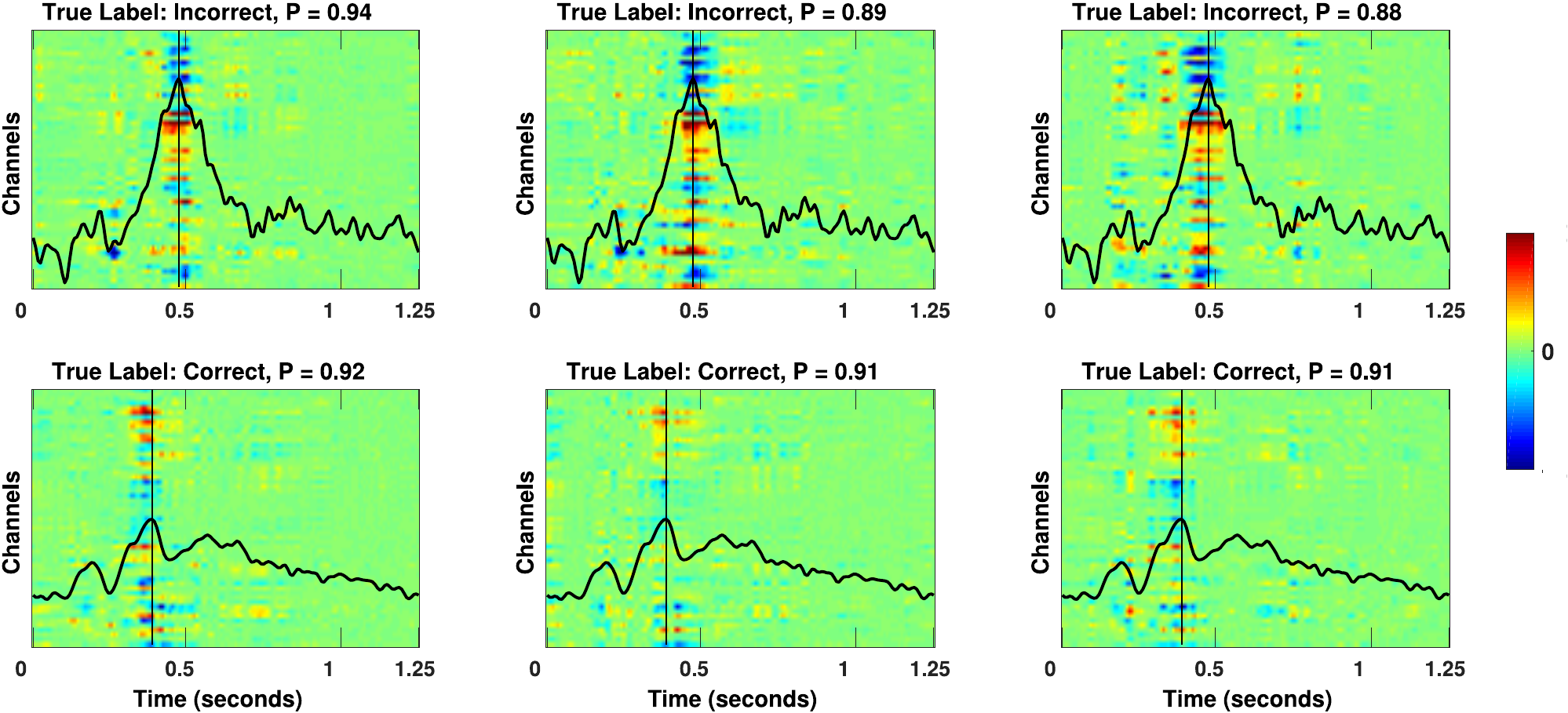}
	\caption{Single-trial EEG feature relevance for a cross-subject trained EEGNet-4,2 model, using DeepLIFT, for the one test subject of the ERN dataset. (Top Row) Feature relevances for three correctly predicted trials of incorrect feedback, along with its predicted probability $P$. (Bottom Row) Same as the top row but for three correctly predicted trials of correct feedback. The black line denotes the average ERP, calculated at channel Cz, for incorrect feedback trials (top row) and for correct feedback trials (bottom row). The thin vertical line denotes the positive peak of the average ERP waveform. Here we see feature relevances coincide strongly with the positive peak of the average ERP waveform for each trial. We also see the positive peak occurring slightly earlier for correct feedback trials versus incorrect feedback trials, consistent with the results in \cite{Margaux2012}.  }
	\label{ERN-DeepLIFT}
\end{figure*}

Figure \ref{ERN-DeepLIFT} shows an additional example of using DeepLIFT to analyze feature relevance for a cross-subject trained EEGNet-4,2 model for one test subject of the ERN dataset. Margaux et. al. \cite{Margaux2012} previously noted that the average ERP for correct feedback trials has an earlier peak positive potential, corresponding to approximately 325 ms, whereas the positive peak potential for incorrect trials occurs slightly later, approximately 475 ms. Here we see the same temporal difference in the timing of the peak positive potential for incorrect feedback trials (vertical line in top row of Figure \ref{ERN-DeepLIFT}) and correct feedback trials (vertical line in bottom row of Figure \ref{ERN-DeepLIFT}). We also see the DeepLIFT feature relevances align very closely to that of the peak positive potential for both classes, suggesting that the network has focused on the peak positive potential as the relevant feature for ERN classification. This finding supports results previously reported in \cite{Margaux2012}, where they showed a strong positive correlation between the amplitude of the peak positive potential and the accuracy of error detection. 

\section{Discussion}\label{discussion}

In this work we proposed \textit{EEGNet}, a compact convolutional neural network for EEG-based BCIs that can generalize across different BCI paradigms in the presence of limited data and can produce interpretable features. We evaluated EEGNet against the state-of-the-art approach for both ERP and Oscillatory-based BCIs across four EEG datasets: P300 visual-evoked potentials, Error-Related Negativity (ERN), Movement-Related Cortical Potentials (MRCP) and Sensory Motor Rhythms (SMR). To the best of our knowledge, this represents the first work that has validated the use of a single network architecture across multiple BCI datasets, each with their own feature characteristics and data set sizes. Our work introduced the use of Depthwise and Separable Convolutions \cite{Chollet16a} for EEG signal classification, and showed that they can be used to construct an EEG-specific model which encapsulates well-known EEG feature extraction concepts. Finally, through the use of feature visualization and ablation analysis, we show that neurophysiologically interpretable features can be extracted from the EEGNet model, providing further validation and evidence that the network performance is not being driven by noise or artifact signals in the data. This last finding is particularly important, as it is a critical component to understanding the validity and robustness of CNN model architectures not just for EEG \cite{Sturm2016, Schirrmeister2017b}, but for CNN architectures in general \cite{Zeiler2014, SimonyanZ14a, Montavon2018}.

The learning capacity of CNNs comes in part from their ability to automatically extract intricate feature representations from raw data. However, since the features are not hand-designed by human engineers, understanding the meaning of those features poses a significant challenge in producing interpretable models \cite{Nguyen2014}. This is especially true when CNNs are used for the analysis of EEG data where features from neural signals are often non-stationary and corrupted by noise artifacts \cite{Waytowich2011, Lawhern2012}. In this study, we illustrated three different approaches for visualizing the features learned by EEGNet: (1) analyzing spatial filter outputs, averaged over trials, on the P300 dataset, (2) visualizing the convolutional kernel weights on the SMR dataset and comparing them to the weights learned by FBCSP, and (3) performing single-trial relevance analysis on the MRCP and SMR datasets. For the ERN dataset we compared single-trial feature relevances to averaged ERPs and showed that relevant features coincided with the peak of the positive potential for correct and incorrect feedback trials, which has been shown in previous literature to be positively correlated to classifier performance \cite{Margaux2012}. In addition, we conducted a feature ablation study to understand the impact of a classification decision on the presence or absence of a particular feature on the P300 dataset. In each of these analyses, we showed that EEGNet was capable of extracting interpretable features that generally corresponded to known neurophysiological phenomena. These results suggest that the classification performances we observed were not due to artifact or noise sources in the data. 

Our results showed that the spatial filters learned by EEGNet for temporal kernels around 12Hz were significantly correlated to the spatial filters learned by FBCSP in the 8-12Hz filter bank for the SMR dataset. This is interesting to note, as the optimization criterion for CSP (optimal variance separation) is different than the optimization criterion for EEGNet (minimum cross-entropy loss). Because of this, it is not guaranteed that the learned filters from these methods would be comparable. It was encouraging to see that many of the filters did in fact overlap (up to a sign ambiguity), suggesting that EEGNet is learning a similar feature representation to that of FBCSP. This analysis is directly enabled by EEGNet's use of depthwise convolutions to tie spatial filters directly to a temporal filter, an aspect that is unique to this model.

Generally speaking, the classification performance of DeepConvNet and EEGNet were similar across all cross-subject analyses, whereas DeepConvNet performance was lower across nearly all within-subject analyses (with the exception of P300). One possible explanation for this discrepancy is the amount of training data used to train the model; in cross-subject analyses the training set sizes were about 10-15 times larger than that of within-subject analyses. This suggests that DeepConvNet is more data-intensive compared to EEGNet, an unsurprising result given that the model size of DeepConvNet is two orders of magnitude larger than EEGNet (see Table \ref{model-size-comparison}). We believe this intuition is consistent with the findings originally reported by the developers of DeepConvNet \cite{Schirrmeister2017b}, where they state that a training data augmentation strategy was needed to obtain good classification performance on the SMR dataset. In contrast to their work, we show that EEGNet performed well across all tested datasets without the need for data augmentation, making the model simpler to use in practice.

In general we found that, both in within- and cross-subject analyses, that ShallowConvNet tended to perform worse on the ERP BCI datasets than on the oscillatory BCI dataset (SMR), while the opposite behavior was observed with DeepConvNet. We believe this is due to the fact that the ShallowConvNet architecture was designed specifically to extract log bandpower features; in situations where the dominant feature is signal amplitude (as is the case in many ERP BCIs), ShallowConvNet performance tended to suffer. The opposite situation occurred with DeepConvNet; as its architecture was not designed to extract frequency features, its performance was lower in situations where frequency power is the dominant feature. In contrast, we found that EEGNet performed just as well as ShallowConvNet in SMR classification and just as well as DeepConvNet in ERP classification (and outperforming in the case of within-subject MRCP, ERN and SMR classifications), suggesting that EEGNet is robust enough to learn a wide variety of features over a range of BCI tasks. 

The severe underperformance of ShallowConvNet on within-subject MRCP classification was unexpected, given the similarity in neural responses between the MRCP and SMR, and the fact that ShallowConvNet performed well on SMR. This discrepancy in performance is not due to the amount of training data used, as within-subject MRCP classification has approximately 700 training trials, evenly split among left and right finger movements, whereas the SMR dataset has only 192 training trials, evenly split among four classes. In addition, we did not observe large deviations in ShallowConvNet performance on the other datasets (P300 and ERN). In fact, ShallowConvNet performed fairly well on within-subject ERN classification, even though this dataset is the smallest among all datasets used in this study (only having 170 training trials total). Determining the underlying source of this phenomena will be explored in future research.

Deep Learning models for EEG generally employ one of three input styles, depending on their targeted application: (1) the EEG signal of all available channels, (2) a transformed EEG signal (generally a time-frequency decomposition) of all available channels \cite{Bashivan2015} or (3) a transformed EEG signal of a subset of channels \cite{Tabar2017}. Models that fall in (2) generally see a significant increase in data dimensionality, thus requiring either more data or more model regularization (or both) to learn an effective feature representation. This introduces more hyperparameters that must be learned, increasing the potential variability in model performance due to hyperparameter misspecification. Models that fall in (3) generally require \textit{a priori} knowledge about the channels to select. For example, the model proposed in \cite{Tabar2017} uses the time-frequency decomposition of channels Cz, C3 and C4 as the inputs for a motor imagery classification task. This channel selection is intentional, given the fact that neural responses to motor actions (the sensory motor rhythm) are observed strongest at those channels and are easily observed through a time-frequency analysis. Also, by only working with three channels, the authors reduce the significant increase in dimensionality of the data. While this approach works well if the feature of interest is known beforehand, this approach is not guaranteed to work well in other applications where the features are not observed at those channels, limiting the overall utility of this approach. We believe models that fall in (1), such as EEGNet and others \cite{Cecotti2011, Shamwell2016, Cecotti2014}, offer the best tradeoff between input dimensionality and the flexibility to discover relevant features by providing all available channels. This is especially important as BCI technologies evolve into novel application spaces, as the features needed for these future BCIs may not be known beforehand  \cite{Zander2011, Lance2012, Saproo2016, vanerp_braincomputer_2012, Blankertz2010, Gordon2017}.

\section*{Acknowledgments}

This project was sponsored by the U.S. Army Research Laboratory under ARL-H70-HR52, ARL-74A-HRCYB and through Cooperative Agreement Number W911NF-10-2-0022. The views and conclusions contained in this document are those of the authors and should not be interpreted as representing the official policies, either expressed or implied, of the U.S. Government. The U.S. Government is authorized to reproduce and distribute reprints for Government purposes notwithstanding any copyright notation herein.

\section*{Conflict of Interest Statement}
The authors declare that the research was conducted in the absence of any commercial or financial relationships that could be construed as a potential conflict of interest.

\clearpage
\newpage

\section{Appendix}

\subsection{DeepConvNet and ShallowConvNet architectures}

The DeepConvNet and ShallowConvNet architectures are given in Tables \ref{deepconvnet-model} and \ref{shallowconvnet-model}, respectively. The DeepConvNet was designed to be a general-purpose architecture that is not restricted to specific feature types, whereas ShallowConvNet is designed specifically for oscillatory signal classification.

\begin{table*}[h!]
	\centering
	\begin{adjustbox}{width=1\textwidth}
		
		\def\arraystretch{1.25}
		\begin{tabular}{l|lllllll}
			\textbf{Layer} & \textbf{\# filters} & \textbf{size}    & \textbf{\# params}  &  \textbf{Activation} & \textbf{Options} \\ \hline
			
			Input               	&         &  (C, T)    &                     &                     &               \\
			Reshape             	&         &  (1, C, T) &                     &                     &               \\
			Conv2D              	& 25      & (1, 5)     & 150                 & Linear               & mode = valid, max norm = 2          \\
			Conv2D     	            & 25      & (C, 1)     & 25 * 25 * C + 25    & Linear               & mode = valid, max norm = 2         \\
			BatchNorm           	&         &            & 2 * 25              &                     & epsilon = 1e-05, momentum = 0.1              \\
			Activation          	&         &            &                     & ELU                 &               \\
			MaxPool2D           	&         & (1, 2)     &                     &                     &               \\
			Dropout             	&         &            &                     &                     & p = 0.5       \\
			
			Conv2D     	            & 50      & (1, 5)     & 25 * 50 * C + 50    & Linear              & mode = valid, max norm = 2         \\
			BatchNorm           	&         &            & 2 * 50              &                     & epsilon = 1e-05, momentum = 0.1              \\
			Activation          	&         &            &                     & ELU                 &               \\
			MaxPool2D           	&         & (1, 2)     &                     &                     &               \\
			Dropout             	&         &            &                     &                     & p = 0.5       \\
			
			Conv2D     	            & 100     & (1, 5)     & 50 * 100 * C + 100  & Linear              & mode = valid, max norm = 2         \\
			BatchNorm           	&         &            & 2 * 100             &                     & epsilon = 1e-05, momentum = 0.1              \\
			Activation          	&         &            &                     & ELU                 &               \\
			MaxPool2D           	&         & (1, 2)     &                     &               \\
			Dropout             	&         &            &                     &                     & p = 0.5       \\
			
			Conv2D     	            & 200     & (1, 5)     & 100 * 200 * C + 200 & Linear              & mode = valid, max norm = 2         \\
			BatchNorm           	&         &            & 2 * 200             &                     & epsilon = 1e-05, momentum = 0.1              \\
			Activation          	&         &            &                     & ELU                 &               \\
			MaxPool2D           	&         & (1, 2)     &                     &                     &               \\
			Dropout             	&         &            &                     &                     & p = 0.5       \\
			Flatten             	&         &                        &                          &                     &               \\
			Dense               	& N       &                        &                          & softmax             &  max norm = 0.5            
		\end{tabular}
	\end{adjustbox}
	\vspace{3mm}
	\caption{DeepConvNet architecture, where $C = $ number of channels,  $T = $ number of time points and $N = $ number of classes, respectively. } \label{deepconvnet-model}
\end{table*}

\begin{table*}[!t]
	\centering
	\begin{adjustbox}{width=1\textwidth}
		
		\def\arraystretch{1.25}
		\begin{tabular}{l|llllllll}
			\textbf{Layer} & \textbf{\# filters} & \textbf{size}    & \textbf{\# params}  &  \textbf{Activation} & \textbf{Options} \\ \hline
			
			Input               	&         & (C, T)                  &               &                     &               \\
			Reshape             	&         & (1, C, T)               &               &                     &               \\
			Conv2D              	& 40      & (1, 13)                 & 560           & Linear              & mode = same, max norm = 2          \\
			Conv2D     	            & 40      & (C, 1)                  & 40 * 40 * C   & Linear              & mode = valid, max norm = 2         \\
			BatchNorm           	&         &                         & 2 * 40        &                     & epsilon = 1e-05, momentum = 0.1              \\
			Activation          	&         &                         &               & square              &               \\
			AveragePool2D       	&         & (1, 35), stride (1, 7) &               &                     &               \\
			Activation          	&         &                         &               & log                 &      \\
			Flatten             	&         &                         &               &                     &               \\
            Dropout             	&         &                         &               &                     & p = 0.5       \\
			Dense               	& N       &                         &               & softmax             &  max norm = 0.5            
		\end{tabular}
	\end{adjustbox}
	\vspace{3mm}
	\caption{ShallowConvNet architecture, where $C = $ number of channels,  $T = $ number of time points and $N = $ number of classes, respectively. Here, the 'square' and 'log' activation functions are given as $f(x) = x^2$ and $f(x) = \log(x)$, respectively. Note that we clip the log function such that the minimum input value is a very small number ($\epsilon = 10e^{-7}$) for numerical stability.  } \label{shallowconvnet-model}
\end{table*}

\clearpage

\footnotesize
\bibliographystyle{IEEEtran}
\bibliography{bibtex_database}

\end{document}